\documentclass[10pt,twocolumn,letterpaper]{article}

\usepackage{cvpr} 
\usepackage{times}
\usepackage{epsfig}
\usepackage{graphicx}
\usepackage{amsmath}
\usepackage{amssymb}
\usepackage{booktabs}
\usepackage{multirow}

\usepackage[pagebackref=true,breaklinks=true,letterpaper=true,colorlinks,bookmarks=false]{hyperref}

\begin{document}

\title{DINO-VPT: Hierarchical Visual Prompt Tuning for Joint Physical-Digital Face Anti-Spoofing}

\author{Pierre Gallin-Martel, Mika Feng, Koichi Ito, and Takafumi Aoki\\
Graduate School of Information Sciences, Tohoku University, Japan\\
{\tt\small \{pierre, mika, ito\}@aoki.ecei.tohoku.ac.jp, aoki@ecei.tohoku.ac.jp}
}

\maketitle
\thispagestyle{empty}

   

\begin{abstract}
  With the increasing diversity of spoofing attacks, there is a growing demand for unified Face Anti-Spoofing (FAS) models capable of detecting both physical and digital threats.
  While existing Vision-Language Models (VLMs) demonstrate high generalization in this context, they heavily rely on complex multimodal fusion and external text encoders.
  In this paper, we propose DINO-VPT, a lightweight, vision-only framework leveraging hierarchical visual prompt tuning.
  By dynamically injecting prompts conditioned on input features via a Prompt Routing Network (PRN), our method effectively disentangles diverse spoofing artifacts without requiring multimodal fusion.
  Evaluations on the UniAttackData benchmark demonstrate that DINO-VPT achieves higher accuracy than state-of-the-art VLM-based methods.
  Our results indicate that a properly structured vision-only architecture can achieve state-of-the-art performance in unified FAS without the need for multimodal supervision.
\end{abstract}

\section{Introduction}

Face recognition (FR) systems have become an indispensable component of recent authentication infrastructures.
However, their widespread adoption has made their vulnerability to spoofing attacks a critical concern.
Consequently, Face Anti-Spoofing (FAS) techniques are vital for ensuring the reliability of FR systems \cite{Handbook-Anti-Spoofing,Handbook-Face-Recognition}.
Early FAS research primarily focused on detecting physical presentation attacks (PAs), such as printed photographs, displays, or masks, achieving high accuracy through deep learning and domain generalization methods \cite{Shao-CVPR-2019,Wang-CVPR-2020,SSAN,Jia-CVPR-2020}.

In recent years, the rapid advancement of generative AI has made digital attack (DA) modalities, including face swapping and deepfakes, easily executable \cite{Dang-CVPR-2020}.
While physical and digital attacks were traditionally treated as distinct tasks, real-world scenarios necessitate addressing both threats simultaneously.
This has led to an increasing demand for Unified Physical-Digital Attack Detection (UAD) \cite{Chen-IJCV-2026}.
However, jointly modeling material-based physical artifacts and rendering-based digital anomalies in a unified feature space is extremely challenging due to the significant intra-class variation inherent in this setting.

Recent UAD studies have proposed leveraging large Vision-Language Models (VLMs) with hierarchical prompt tuning based on language guidance \cite{Liu-arXiv-2025}.
While these multimodal approaches exhibit strong generalization, they incur substantial computational costs and depend heavily on external text encoders, posing challenges for practical deployment in resource-constrained environments.
Moreover, direct fine-tuning of large vision foundation models on UAD data is prone to overfitting, making it difficult to disentangle fine-grained differences between attack types.

Feng et al. \cite{Feng-ICCVW-2025} demonstrated that DINOv2 with registers \cite{Darcet-ICLR-2024} is highly effective as a vision foundation model for capturing the minute visual discrepancies characteristic of spoofing attacks.
However, even with the powerful representation capabilities of DINOv2, a single feature extraction strategy remains insufficient to cover the diverse attack spectrum in UAD.
To address this issue, we propose DINO-VPT, a lightweight vision-only framework built upon DINOv2.
We introduce hierarchical visual prompt tuning without relying on heavy textual supervision.
Specifically, we designed a lightweight Prompt Routing Network (PRN) that dynamically controls prompt injection based on input-level features.
This mechanism enables the efficient separation of attack types across coarse to fine-grained levels and the acquisition of sophisticated feature representations without the need for multimodal fusion.

Through extensive experiments, we demonstrate the effectiveness of the proposed DINO-VPT.
Evaluation on the UniAttackData benchmark \cite{Chen-IJCV-2026} shows that DINO-VPT outperforms state-of-the-art VLM-based methods despite their massive computational requirements.
Furthermore, on the standard cross-domain MICO protocol, our method maintains generalization performance comparable to traditional image-based state-of-the-art methods.
These results indicate that a properly structured vision-only design can provide a high-accuracy and efficient solution for UAD without the need for multimodal supervision.

\section{Related Work}

The focus of FAS research has transitioned from detecting specific attacks to generalizing across unseen domains and diverse attack modalities.
This section provides an overview of vision foundation models for FAS, multimodal approaches, and the emerging challenge of unified physical-digital attack detection.

\subsection{Vision Foundation Models for FAS}

Traditional FAS research has primarily utilized CNN-based methods to capture local texture artifacts, such as CDCN \cite{Yu-CVPR-2020}, PatchNet \cite{Wang-CVPR-2022}, ResNet50 \cite{He-CVPR-2016}, and models incorporating auxiliary supervision \cite{Liu-CVPR-2018}.
However, because CNNs have a limited receptive field, they are often insufficient for modeling the global dependencies required to represent complex spoofing patterns.
Recently, Vision Transformers (ViTs) \cite{Dosovitskiy-ICLR-2021} have achieved success in FAS due to their ability to capture long-range interactions via self-attention mechanisms.
Feng et al. \cite{Feng-CVPRW-2025} demonstrated that intermediate ViT features retain robust spoofing representations.
Building on this, Feng et al. \cite{Feng-ICCVW-2025} demonstrated that DINOv2 with registers \cite{Darcet-ICLR-2024} is highly effective for capturing the minute visual discrepancies characteristic of spoofing attacks.
Nevertheless, these vision-only methods rely on static feature extraction, which limits their adaptability to domain shifts.

\subsection{Multimodal FAS and Prompt-Based Tuning}

To enhance domain generalization, multimodal approaches leveraging vision-language models (VLMs) such as CLIP \cite{Radford-ICML-2021} have been proposed.
Frameworks like FLIP \cite{Srivatsan-ICCV-2023} and CFPL \cite{Liu-ICCV-2023} improve robustness against unknown attacks by introducing semantic language guidance.
These methods have established state-of-the-art (SOTA) performance on the standard MICO cross-domain protocol, which involves evaluation across diverse datasets including MSU-MFSD \cite{MSU-MFSD}, CASIA-FASD \cite{CASIA-FASD}, IDIAP Replay-Attack \cite{IDIAP}, and OULU-NPU \cite{Boulkenafet-FG-2017}.
Additionally, parameter-efficient fine-tuning via hierarchical prompt tuning, such as HIPTune \cite{Liu-arXiv-2025}, has been proposed to effectively adapt VLMs for FAS tasks.
However, these multimodal frameworks introduce substantial computational overhead and depend heavily on external text encoders, limiting their practical deployment.

\subsection{Unified Physical-Digital Attack Detection (UAD)}

The primary focus of this work is Unified Physical-Digital Attack Detection (UAD), which was recently proposed by Chen et al. \cite{Chen-IJCV-2026} as a non-trivial and emerging task in FAS.
While conventional FAS methods address physical presentation attacks (PAs) and digital attacks (DAs) independently, the UAD paradigm necessitates a unified framework capable of detecting both attack modalities simultaneously.
Extracting discriminative features that capture both material-based physical cues and rendering-based digital artifacts is challenging due to the significant intra-class variation between these domains.
Existing works such as MTFace \cite{He-CVPRW-2024}, SUEDE \cite{Xie-ICME-2025}, and MoAE-CR \cite{Chen-AAAI-2025} have pioneered this direction.
However, the design of vision-only architectures that can effectively bridge the performance gap with VLM-based models remains an open challenge. 
This paper explores the potential of a vision-only framework for robust UAD.

\begin{figure*}[t]
  \centering
  \includegraphics[width=\linewidth]{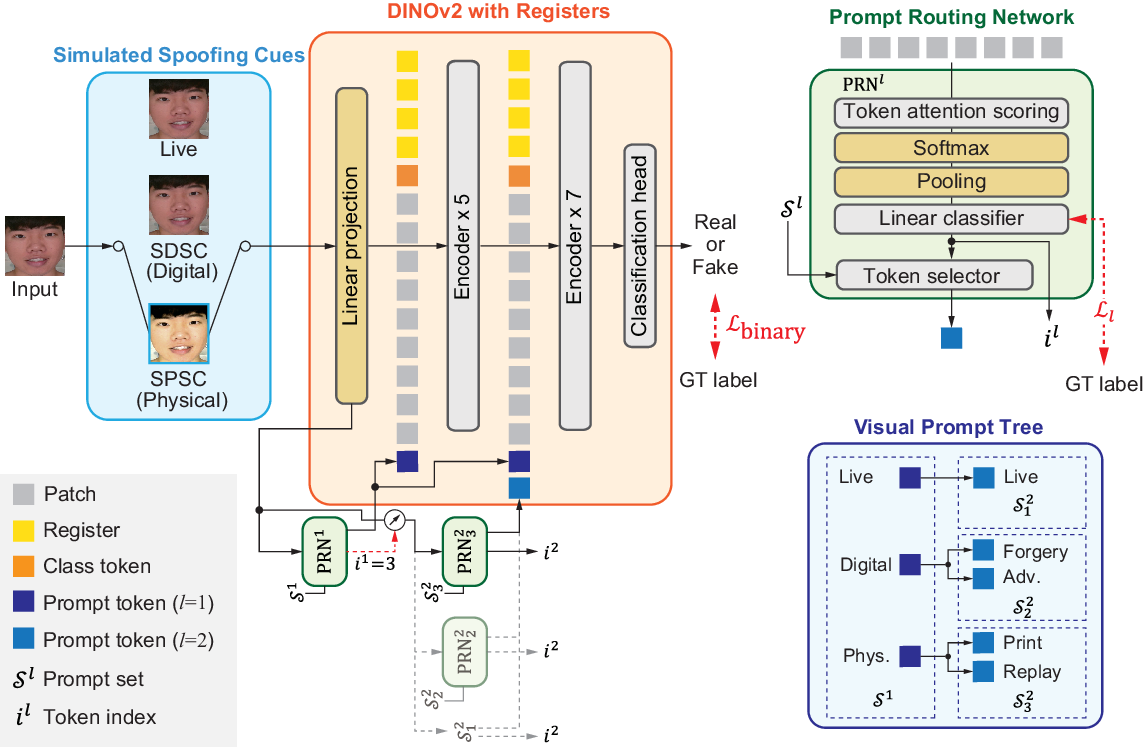}
  \caption{Overview of the proposed DINO-VPT framework. During training, an input image first passes through the Simulated Spoofing Cues (SSC) module, which stochastically applies domain-specific augmentations (Physical or Digital) for dynamic relabeling. The image is then processed by a DINOv2 backbone with registers, where learnable visual prompts are injected at multiple levels. These prompts are dynamically selected by Prompt Routing Networks (PRNs) operating directly on initial patch embeddings. Guided by the Visual Prompt Tree (VP-Tree), each PRN functions as a conditional multiplexer: it retrieves a specific prompt from the prompt set $\mathcal{S}^l$ for injection, while simultaneously outputting an index $i^l$ to conditionally activate the highly specialized routing module at the next hierarchy level. This mechanism enables progressive adaptation from coarse attack modalities to fine-grained subtypes.}
  \label{fig:method}
\end{figure*}

\section{Proposed Method: DINO-VPT}

Our proposed DINO-VPT introduces an efficient, vision-only framework tailored for UAD.
The core idea is to adapt a powerful vision foundation model to heterogeneous spoofing cues using a hierarchical, dynamic prompt injection mechanism.
By organizing task-specific knowledge into a structured Visual Prompt Tree (VP-Tree) and dynamically selecting these prompts via lightweight routing networks, the framework achieves progressive specialization from coarse attack modalities to fine-grained subtypes. 
Unlike prior multimodal formulations, our approach operates in the visual domain, enabling a simpler and more efficient design while supporting robust generalization.

The overall pipeline is illustrated in Fig. \ref{fig:method}.
During training, an input image first passes through the Simulated Spoofing Cues (SSC) module, which stochastically applies domain-specific augmentations and dynamic relabeling to ensure comprehensive coverage of the hierarchical label space.
The image is then converted into a sequence of patch embeddings via a linear projection layer and processed by a pretrained DINOv2 ViT-B/14 backbone with registers \cite{Darcet-ICLR-2024}.
To adapt these representations to spoof-specific cues, learnable visual prompts are injected at multiple depths of the transformer, enabling layer-wise modulation of features.
This injection is dynamically governed by PRNs operating directly on the initial input-level features.
At each hierarchy level $l$, the active PRN functions as a multiplexer: it selects a specific prompt token from the VP-Tree for injection, while simultaneously outputting a token index ($i^l$) to conditionally activate the appropriate PRN at the subsequent depth.
The final representation, obtained from the class token of the last encoder block, is passed to a classification head to predict the liveness label.
The entire framework, including the routing decisions and binary classification, is jointly optimized end-to-end. 
Specifically, the backbone is initialized with pretrained weights and fully fine-tuned, while prompt embeddings, PRNs, and the classification head are learned from scratch.

The subsequent sections detail the core components of our architecture: the foundation backbone (Sec. \ref{sec:backbone}), the structure of VP-Tree (Sec. \ref{sec:vp-tree}), PRN (Sec. \ref{sec:prn}), SSC (Sec. \ref{sec:ssc}), and the loss functions (Sec. \ref{sec:loss}).

\subsection{BackBone: DINOv2 with Registers}
\label{sec:backbone}

UAD requires highly generalized and robust image features to capture diverse spoofing cues, ranging from subtle physical artifacts (e.g., moir\'e patterns) to sophisticated digital manipulations.
Therefore, we adopt DINOv2 with registers \cite{Darcet-ICLR-2024}, specifically the ViT-B/14 architecture, as the backbone of our framework. 
Unlike traditional FAS models trained from scratch on limited domains, DINOv2 provides general-purpose visual representations that encode both high-level global semantics and fine-grained spatial details necessary for identifying heterogeneous attack patterns.
The efficacy of this architecture in the FAS domain has been recently established by Feng et al. \cite{Feng-ICCVW-2025}, who demonstrated its superiority over standard supervised backbones in cross-domain spoofing detection. 
Furthermore, the inclusion of register tokens is particularly essential for our framework. 
Standard ViT models often exhibit high-norm outlier tokens in non-informative regions, which can disrupt the feature representation and interfere with the extraction of subtle spoofing cues. 
Registers effectively absorb these uninformative signals, leading to more stable and concentrated attention maps. 
Consequently, this stability enhances our dynamic routing mechanism, ensuring that the patch-level features processed by PRNs reflect task-relevant information rather than background noise.


\subsection{Hierarchical Visual Prompt Injection (VP-Tree)}
\label{sec:vp-tree}

Expanding on prior hierarchical prompt tuning approaches such as HiPTune \cite{Liu-arXiv-2025}, we organize task-specific visual prompts within a structured representation referred to as the VP-Tree.
We adapt and simplify its architecture to align with the unified attack taxonomy defined in UAD benchmarks.
In this tree structure, each node is associated with a single learnable prompt token, and each level $l$ corresponds to progressively finer-grained attack subtypes.
The root level ($l=1$) contains coarse categories representing ``Live,'' ``Digital,'' and ``Physical'' modalities ($\mathcal{S}^1$).
The subsequent level ($l=2$) refines these into specific attack subtypes ($\mathcal{S}^2_k$): ``Digital'' attacks branch into ``Forgery'' and ``Adversarial,'' while ``Physical'' attacks branch into ``Print'' and ``Replay.''
To maintain a consistent tensor structure during routing, a dedicated identity prompt is also assigned to the ``Live'' class at $l=2$.
These visual prompts act as task-specific conditioning signals and are injected at different depths of the transformer sequence, following the hierarchical progression of the VP-Tree.
Specifically, the selected prompt at $l=1$ is injected immediately after the linear projection layer (prior to the first encoder block), whereas the $l=2$ prompt is introduced before the 5th transformer block.
This depth-wise injection encourages layer specialization at different levels of granularity while propagating task-relevant information to deeper layers.
Unlike traditional prompt tuning that freezes the foundation model, we adopt this paradigm in conjunction with full backbone fine-tuning using a smaller learning rate for the pretrained parameters.
This joint optimization enables the backbone to adaptively modulate the influence of the injected prompts without losing its inherent generalization capabilities.

\subsection{Prompt Routing Network (PRN)}
\label{sec:prn}

To dynamically select the most appropriate prompts from the VP-Tree, we introduce the PRN. 
As illustrated in Fig. \ref{fig:method}, each PRN operates directly on the initial patch embeddings obtained from the linear projection layer.
Unlike class token based routing strategies \cite{Wang-CVPR-2022-prompt,Wang-ECCV-2022} that rely on intermediate layers for prompt selection, our approach leverages rich spatial information directly from the input level. 
This maintains a consistent input space and avoids circular dependencies between prompt selection and feature adaptation. 
The PRN employs a token attention scoring mechanism followed by a softmax operation. 
These normalized scores are used to perform a weighted pooling of the patch features, yielding a global representation that is passed to a linear classifier.
Each PRN functions as a conditional multiplexer with a dual-output mechanism. 
At a given level $l$, the linear classifier predicts a probability distribution over the available child nodes, determining the winning token index $i^l$. 
This index $i^l$ serves two distinct purposes. 
First, it is passed to a token selector, which retrieves the corresponding visual prompt from the prompt set $\mathcal{S}^l$ for injection into the transformer sequence. 
For $l=1$ and $l=2$, prompts are empirically injected before the 1st and 6th encoder blocks to optimize feature modulation as shown in Fig. \ref{fig:method}.
Second, $i^l$ acts as a control signal to activate the specific PRN module associated with this index at the subsequent level.
This conditional activation enforces consistency across the VP-Tree. 
For instance, the root router ($\text{PRN}^1$) evaluates the coarse categories and outputs an index $i^1$. 
If the network predicts the ``Physical'' class (e.g., $i^1 = 3$), only the corresponding downstream router ($\text{PRN}^2_3$) is evaluated at the next depth ($l=2$), completely bypassing the digital branch. 
This dynamic routing not only prevents incompatible prompts from being injected simultaneously but also simplifies the decision space for deeper routers by restricting them to highly specialized classification tasks.

\begin{figure}[t]
  \centering
  \includegraphics[width=\linewidth]{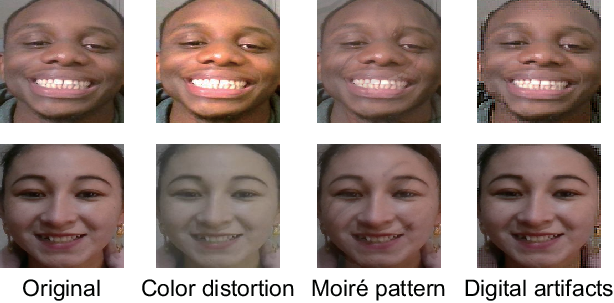}
  \caption{Visual examples of dynamic relabeling applied by the Simulated Spoofing Cues (SSC) module. From left to right, the columns display: the original Live image, Color Distortion (SPSC for Print simulation), Moir\'e Pattern (SPSC for Replay simulation), and geometric inconsistencies (SDSC for Digital simulation). These stochastically applied transformations ensure comprehensive coverage of the hierarchical attack taxonomy defined in our VP-Tree.}
  \label{fig:SSC}
\end{figure}

\subsection{Simulated Spoofing Cues (SSC)}
\label{sec:ssc}

To ensure meaningful coverage of all attack subtypes within the VP-Tree, we incorporate SSC into the data augmentation pipeline. Rather than serving as a conventional augmentation strategy, SSC acts as a dynamic relabeling mechanism. As shown in Fig. \ref{fig:method}, a subset of live images is stochastically transformed into synthetic physical or digital spoof samples during training. These samples are relabeled as ``Spoof,'' with subtype annotations updated to match the corresponding VP-Tree routing paths.
SSC employs Simulated Physical Spoofing Cues (SPSC) and Simulated Digital Spoofing Cues (SDSC) \cite{He-CVPRW-2024}. SPSC simulates physical attacks through appearance degradations, approximating print attacks with severe color perturbations and replay attacks with moir'e-like patterns. Conversely, SDSC simulates digital attacks by introducing inconsistencies between facial appearance and geometry via spatial transformations of the input image and its corresponding depth map, producing artifacts typical of face swapping and generative pipelines. Adversarial attacks are not simulated due to the lack of a reliable appearance-based approximation strategy.
By exposing the model to these simulated variations, SSC alleviates class imbalance within the tree branches and improves the robustness of both routing decisions and feature representations. Examples are shown in Fig. \ref{fig:SSC}.



\subsection{Loss Function}
\label{sec:loss}

To jointly optimize the classification head and the hierarchical routing modules, the entire DINO-VPT framework is trained end-to-end using a combined objective function.
The total loss is formulated as:
\begin{equation}
  \mathcal{L}_{\text{total}} = \lambda_{\text{bin}} \mathcal{L}_{\text{binary}} + \lambda_{1} \mathcal{L}_{\text{route}}^1 + \lambda_{2} \mathcal{L}_{\text{route}}^2,
\end{equation}
where $\mathcal{L}_{\text{binary}}$ is the objective for the final binary (live vs. spoof) classification, and $\mathcal{L}_{\text{route}}^1, \mathcal{L}_{\text{route}}^2$ denote the auxiliary routing losses at levels $l=1$ and $l=2$ of the VP-Tree, respectively.
The hyperparameters $\lambda_{\text{bin}}$, $\lambda_{1}$, and $\lambda_{2}$ balance the contribution of each term.

\paragraph{Binary Classification Loss.}

To address varying difficulty levels across heterogeneous spoofing attacks, the final classification head is supervised using Focal Loss calculated by
\begin{equation}
  \mathcal{L}_{\text{binary}} = - \alpha \sum_{c \in \{0, 1\}} (1 - p_c)^{\gamma} \, y_c \log p_c,
\end{equation}
where $p_c$ is the predicted probability for class $c$, $y_c$ is the one-hot ground-truth label, and $\gamma$ is the focusing parameter.
While $\alpha$ typically controls class weighting, we apply it uniformly in our implementation, as severe class imbalance is inherently mitigated through weighted random sampling during training and the dynamic relabeling of the SSC module.

\paragraph{Routing Losses.}

To ensure that the PRNs learn a meaningful partitioning of the spoofing cues, each router is trained using a standard cross-entropy loss over its specific subtype labels, rather than simple binary supervision. 
Specifically, $\mathcal{L}_{\text{route}}^1$ optimizes the root router ($\text{PRN}^1$) to classify the coarse modalities defined in $\mathcal{S}^1$ (``Live,'' ``Physical,'' ``Digital''). 
Similarly, $\mathcal{L}_{\text{route}}^2$ optimizes the conditionally activated child router at the subsequent depth ($l=2$) for fine-grained subtype classification defined in $\mathcal{S}^2_k$ (e.g., ``Print'' vs. ``Replay'' for the physical branch, or ``Forgery'' vs. ``Adversarial'' for the digital branch). 
These auxiliary routing losses are computed exclusively for the active branches, strictly enforcing the conditional execution flow dictated by the VP-Tree.

\section{Experiments and Discussion}

This section presents a comprehensive evaluation of the proposed DINO-VPT framework.
We first describe the details of the datasets, evaluation metrics, and implementation settings. Subsequently, we evaluate our method against state-of-the-art approaches on both the unified face anti-spoofing benchmark (UniAttackData) and the cross-dataset benchmark (MICO).
Finally, we analyze the computational efficiency of our approach and conduct extensive ablation studies to validate the core components.

\begin{table*}[t]
  \centering
  \caption{Experimental results on Protocols 1 and 2 of UniAttackData. Protocol 2 reports the average of results from 2.1 and 2.2 (mean $\pm$ std). For MTFace and our proposed method, the specific results for P2.1 and P2.2 are provided in parentheses below the average as (P2.1, P2.2). The best and second-best results are highlighted in \textbf{bold} and \underline{underline}, respectively. Metrics marked as ``---'' indicate values not reported in the original papers. All metrics are reported as percentages (\%).}
  \label{tbl:UAD}
  \begin{tabular}{@{}lcccccc@{}}
    \toprule
    \multirow{2}{*}{Method} & \multicolumn{3}{c}{P1} & \multicolumn{3}{c}{P2} \\
    \cmidrule(lr){2-4} \cmidrule(lr){5-7}
    & ACER$\downarrow$ & AUC$\uparrow$ & ACC$\uparrow$
    & ACER$\downarrow$ & AUC$\uparrow$ & ACC$\uparrow$ \\
    \midrule
    UAD+ (baseline) \cite{Chen-IJCV-2026} & 0.36 & 99.97 & \underline{99.53} & 9.77$\pm$6.05 & 96.31$\pm$2.91 & 75.76$\pm$20.73 \\
    ResNet50 \cite{He-CVPR-2016} & 1.35 & 99.79 & 98.83 & 34.60$\pm$5.31 & 87.89$\pm$6.11 & 53.69$\pm$6.39 \\
    ViT-B/16 \cite{Dosovitskiy-ICLR-2021} & 5.92 & 97.00 & 92.29 & 33.69$\pm$9.33 & 83.77$\pm$2.35 & 52.43$\pm$25.88 \\
    DINOv2 w/ Reg \cite{Darcet-ICLR-2024} & 3.40 & 99.54 & 97.00 & 18.55$\pm$0.11 & 89.32$\pm$0.02 & 82.83$\pm$1.14 \\
    Auxiliary \cite{Liu-CVPR-2018} & 1.13 & 99.82 & 98.68 & 42.98$\pm$6.77 & 76.27$\pm$12.06 & 37.71$\pm$26.45 \\
    CDCN \cite{Yu-CVPR-2020} & 1.40 & 99.52 & 98.57 & 34.33$\pm$0.66 & 77.46$\pm$17.56 & 53.10$\pm$12.70 \\
    FFD \cite{Dang-CVPR-2020} & 2.01 & 99.57 & 97.97 & 44.20$\pm$1.32 & 80.97$\pm$2.86 & 40.43$\pm$14.88 \\
    MoAE-CR \cite{Chen-AAAI-2025} & 0.37 & 99.97 & 99.47 & 15.13$\pm$12.10 & 92.09$\pm$7.11 & 85.41$\pm$6.85 \\
    SUEDE \cite{Xie-ICME-2025} & 0.36 & 99.70 & 99.50 & 11.99$\pm$7.79 & 91.49$\pm$4.99 & \underline{88.99$\pm$6.96} \\
    HIPTune \cite{Liu-arXiv-2025} & 0.33 & \textbf{99.99} & --- & 10.38 & \underline{97.82} & --- \\
    \midrule
    \multirow{2}{*}{MTFace \cite{He-CVPRW-2024}} 
    & \underline{0.20} & --- & --- 
    & \underline{1.49}$\pm$0.17 & --- & --- \\
    & & & 
    & (1.32, 1.65) & & \\
    
    \midrule
    
    \multirow{2}{*}{DINO-VPT (ours)} 
    & \textbf{0.07} & \textbf{99.99} & \textbf{99.91} 
    & \textbf{0.63$\pm$0.18} & \textbf{99.75$\pm$0.20} & \textbf{99.33$\pm$0.23} \\
    & & & 
    & (0.45, 0.81) & (99.55, 99.95) & (99.56, 99.10) \\
    \bottomrule
  \end{tabular}
\end{table*}

\subsection{Datasets and Evaluation Metrics}

To evaluate the proposed framework, we conduct experiments on two benchmarks: UniAttackData \cite{Chen-IJCV-2026} for unified physical-digital attack detection, and the MICO benchmark for standard cross-dataset physical presentation attack detection.

\paragraph{UniAttackData.}

UniAttackData is the largest unified face anti-spoofing dataset, comprising 28,706 images from 1,800 subjects. It contains 1,800 live faces, 5,400 physical attacks (PAs), and 21,506 digital attacks (DAs), including adversarial examples. A key feature is its strict identity consistency, where complete attack types are available for every identity across both physical and digital domains. This prevents models from relying on identity-specific cues instead of spoofing artifacts.
Following the official protocol, we evaluate under two settings. Protocol 1 (P1) measures intra-domain performance, with training, validation, and evaluation sets sharing all attack modalities but differing in subject identities. Protocol 2 (P2) evaluates cross-domain generalization using a leave-one-type-out strategy. In P2.1, models are trained on digital and adversarial attacks and evaluated on unseen physical attacks, while P2.2 trains on physical attacks and evaluates on unseen digital and adversarial attacks. We also compare against the benchmark baseline, UniAttackDetection+ (UAD+) \cite{Chen-IJCV-2026}.

\paragraph{MICO Benchmark.}

To verify that our model retains robust generalization against presentation attacks, we use the widely adopted MICO protocol.
This involves four standard datasets: MSU-MFSD \cite{MSU-MFSD} (M), IDIAP Replay-Attack \cite{IDIAP} (I), CASIA-FASD \cite{CASIA-FASD} (C), and OULU-NPU \cite{Boulkenafet-FG-2017} (O).
Following the leave-one-dataset-out strategy, the model is trained on three datasets and tested on the remaining unseen dataset.

\paragraph{Evaluation Metrics.}

For UniAttackData, we follow the ISO/IEC 30107-3 standards and report the Average Classification Error Rate (ACER) as the primary ranking metric.
ACER is defined as the mean of the Attack Presentation Classification Error Rate (APCER) and the Bona Fide Presentation Classification Error Rate (BPCER).
The determination of ACER on the test set relies on the threshold established at the Equal Error Rate (EER) of the evaluation set.
We also report classification accuracy (ACC) and the Area Under the ROC Curve (AUC) to provide complementary perspectives.
For the MICO benchmark, performance is measured using the Half Total Error Rate (HTER) and AUC.

\begin{table*}[t]
  \centering
  \caption{Experimental results on the MICO benchmark [\%]. The best and second-best results among vision-only methods are highlighted in \textbf{bold} and \underline{underline}, respectively. VLM-based multimodal methods are listed at the bottom for reference.}
  \label{tbl:results_MICO}
  \setlength{\tabcolsep}{6pt}
  \begin{tabular}{@{}lcccccccccc@{}} 
    \toprule
    \multirow{2}{*}{Method} &
      \multicolumn{2}{c}{CIO$\rightarrow$M} &
      \multicolumn{2}{c}{OMI$\rightarrow$C} &
      \multicolumn{2}{c}{OCM$\rightarrow$I} &
      \multicolumn{2}{c}{ICM$\rightarrow$O} &
      \multicolumn{2}{c}{Avg.} \\
      \cmidrule(lr){2-3} \cmidrule(lr){4-5} \cmidrule(lr){6-7} \cmidrule(lr){8-9} \cmidrule(lr){10-11}
     & HTER$\downarrow$ & AUC$\uparrow$ & HTER$\downarrow$ & AUC$\uparrow$ & HTER$\downarrow$ & AUC$\uparrow$ & HTER$\downarrow$ & AUC$\uparrow$ & HTER$\downarrow$ & AUC$\uparrow$ \\
    \midrule
    DRDG \cite{Liu-IJCAI-2021} & 12.43 & 95.81 & 19.05 & 88.79 & 15.56 & 91.79 & 15.63 & 91.75 & 15.67 & 92.04 \\
    ANRL \cite{Liu-ACM-2021} & 10.83 & 96.75 & 17.83 & 89.26 & 16.03 & 91.04 & 15.67 & 91.90 & 15.09 & 92.24 \\
    PatchNet \cite{Wang-CVPR-2022} & 7.10 & 98.46 & 11.33 & 94.58 & 13.40 & 95.67 & 11.82 & 95.07 & 10.91 & 95.95 \\
    DiVT-M \cite{Liao-WACV-2023} & \textbf{2.86} & \textbf{99.14} & 8.67 & 96.92 & \textbf{3.71} & \textbf{99.29} & 13.06 & 94.04 & 7.08 & 97.35 \\
    IADG \cite{Zhou-CVPR-2023} & 5.41 & 98.19 & 8.70 & 96.44 & 10.62 & 94.50 & 8.86 & 97.14 & 8.40 & 96.57 \\
    GAC-FAS \cite{Le-CVPR-2024} & \underline{5.00} & 97.56 & \underline{8.20} & 95.16 & \underline{4.29} & \underline{98.87} & 8.60 & 97.16 & \textbf{6.52} & 97.19 \\
    Li \cite{Li-NN-2024} & 12.92 & 94.33 & 9.26 & \underline{96.98} & 10.87 & 95.46 & 15.13 & 91.43 & 12.05 & 94.55 \\
    Feng \cite{FENG-arXiv-2026} & 8.86 & 96.95 & \textbf{4.49} & \textbf{98.92} & 9.81 & 96.70 & \underline{7.35} & \underline{98.07} & 7.63 & \textbf{97.66} \\
    \midrule
    DINO-VPT (ours) & 9.14 & 96.61 & 10.62 & 96.10 & 12.53 & 95.75 & \textbf{6.56} & \textbf{98.10} & 9.71 & 96.64\\
    \midrule
    FLIP \cite{Srivatsan-ICCV-2023} & 4.95 & 98.11 & 0.54 & 99.98 & 4.25 & 99.07 & 2.31 & 99.63 & 3.01 & 99.20 \\
    CFPL \cite{Liu-ICCV-2023} & 1.43 & 99.28 & 2.56 & 99.10 & 5.43 & 98.41 & 2.50 & 99.42 & 2.98 & 99.05 \\
    I-FAS \cite{Zhang-AAAI-2025} & 0.32 & 99.88 & 0.04 & 99.99 & 3.22 & 98.48 & 1.74 & 99.66 & 1.33 & 99.50 \\
    \bottomrule
  \end{tabular}
\end{table*}

\begin{table}[t]
  \centering
  \caption{Parameter count for different methods. Trainable parameters are estimated based on reported architectures when not explicitly provided. The lowest parameter counts are highlighted in \textbf{bold}.}
  \label{tbl:TotalParam}
  \begin{tabular}{@{}lcc@{}}
    \toprule
    Method & Trainable [M] & Total [M] \\
    \midrule
    CFPL \cite{Liu-ICCV-2023} & 94 & 157 \\
    FLIP \cite{Srivatsan-ICCV-2023} & 170 & 170 \\
    I-FAS \cite{Zhang-AAAI-2025} & 104 & 3,100 \\
    HiPTune \cite{Liu-arXiv-2025} & \textbf{3.4} & 150 \\
    UAD+ \cite{Chen-IJCV-2026}& 18 & 150 \\
    MTFace \cite{He-CVPRW-2024} & 25.6 & \textbf{25.6} \\
    \midrule
    DINO-VPT & 87.4 & 87.4 \\
    \bottomrule
  \end{tabular}
\end{table}

\subsection{Implementation Details}

DINO-VPT is implemented in PyTorch and optimized end-to-end for 200 epochs using AdamW \cite{AdamW} with early stopping (patience = 20). The ViT-B/14 backbone with registers is initialized from DINOv2 pretrained weights and fine-tuned with a learning rate of $5 \times 10^{-6}$. Prompt embeddings, routing modules (PRNs), and the classification head are randomly initialized and trained with a learning rate of $5 \times 10^{-5}$.

Training uses a batch size of 32. To address class imbalance, we employ a weighted random sampler maintaining a target live sample ratio of 0.3 per batch. This empirically provides an optimal balance for the root routing task ($l=1$), preventing augmented spoof samples from overshadowing real attack distributions. Focal loss parameters are set to $\gamma = 2$ and $\alpha = 0.9$. Loss weights are $\lambda_{\text{bin}} = 1$, $\lambda_{1} = 1$, and $\lambda_{2} = 0.5$, reducing the contribution of deeper routing supervision.

SSC augmentation probabilities are configured per protocol to compensate for missing modalities. For P1, SPSC and SDSC are both applied with probability 0.2. For P2.1 (digital-to-physical), only SPSC is applied with probability 0.6 to synthesize the missing physical domain. For P2.2 (physical-to-digital), only SDSC is applied with probability 0.2, preventing overfitting to forgery artifacts while preserving generalization to unseen digital attacks. Unless otherwise specified, all optimization and architectural hyperparameters are shared across experiments. Protocol-specific augmentation probabilities are adjusted only to reflect the modality imbalance of each protocol.

\subsection{Experimental Results}

In this section, we present the quantitative evaluation of the proposed DINO-VPT framework.
We first compare our method against state-of-the-art approaches on the UniAttackData and MICO benchmarks.
We then analyze the computational efficiency of our model and conclude with comprehensive ablation studies to validate the effectiveness of our core architectural components.

\paragraph{Results on UniAttackData.}

Table \ref{tbl:UAD} summarizes the performance of our method on the UniAttackData benchmark.
For a comprehensive comparison, we include state-of-the-art VLM-based approaches (e.g., UAD+ \cite{Chen-IJCV-2026}, MoAE-CR \cite{Chen-AAAI-2025}, SUEDE \cite{Xie-ICME-2025}, and HiPTune \cite{Liu-arXiv-2025}) that leverage robust vision-language priors, as well as MTFace \cite{He-CVPRW-2024}, a vision-only baseline that utilizes simulated spoofing cues on a ResNet-50 backbone.
Our DINO-VPT framework achieves state-of-the-art performance across both evaluation protocols.
While most approaches perform strongly on Protocol 1 (seen attacks), they suffer significant performance degradation on Protocol 2 due to the severe domain shift introduced by unseen attack modalities.
In contrast, our method attains an average ACER of 0.63\% on P2.
Furthermore, a detailed analysis of P2 demonstrates consistent cross-modality generalization in both directions, achieving 0.45\% ACER for digital-to-physical (P2.1) and 0.81\% ACER for physical-to-digital (P2.2).
This balanced performance demonstrates that structuring attack representations through the hierarchical VP-Tree, combined with dynamic relabeling, captures transferable features across distinct spoofing mechanisms without relying on external textual data.





\paragraph{Cross-Dataset Evaluation on MICO.}

To evaluate generalization in traditional physical-only attack scenarios, we use the MICO benchmark under a leave-one-dataset-out protocol.
As shown in Table \ref{tbl:results_MICO}, our vision-only DINO-VPT achieves performance comparable to state-of-the-art image-encoder-based methods.
While VLM-based approaches (e.g., FLIP, I-FAS) achieve lower error rates, they heavily rely on massive model sizes and external textual supervision.
In contrast, our framework maintains robust generalization against standard physical attacks without requiring cross-modal priors.
This result confirms that optimizing explicitly for unified physical-digital attack detection does not compromise the model's fundamental physical anti-spoofing capabilities.



\paragraph{Computational Efficiency.}

Table \ref{tbl:TotalParam} compares the parameter efficiency of our method against recent state-of-the-art approaches.
Prompt-based multimodal methods (e.g., HiPTune, UAD+) exhibit a very small number of trainable parameters by freezing large vision-language backbones.
However, these models still require approximately 150M to over 3B total parameters, making them computationally heavy during inference.
Conversely, while MTFace is highly compact (25.6M), it suffers from performance degradation on the cross-modality P2 setting.
In contrast, our DINO-VPT operates on a single vision backbone with a total of 87.4M parameters, as the prompt embeddings and routing modules introduce negligible overhead. The resulting model requires 44.94 GFLOPs per inference and achieves an average inference latency of 0.29 ms, measured on an NVIDIA A100 80GB GPU.
This result demonstrates a high performance-efficiency trade-off: achieving the highest UAD accuracy through a compact, end-to-end architecture without relying on massive auxiliary models.






\begin{table*}[t]
  \centering
  \caption{Ablation study on the impact of the hierarchical depth and SSC. ``None'' indicates the absence of PRN modules (equivalent to the standalone DINOv2 with Registers backbone). The specific results for P2.1 and P2.2 are provided in parentheses below the average as (P2.1, P2.2). All metrics are reported as percentages [\%].}
  \label{tbl:Ablation}
  \setlength{\tabcolsep}{6pt}
  \begin{tabular}{@{}cccccccc@{}}
    \toprule
    \multirow{2}{*}{SSC} & \multirow{2}{*}{PRN Depth} 
    & \multicolumn{3}{c}{P1} 
    & \multicolumn{3}{c}{P2} \\
    \cmidrule(lr){3-5} \cmidrule(lr){6-8}
    & & ACER$\downarrow$ & AUC$\uparrow$ & ACC$\uparrow$ 
      & ACER$\downarrow$ & AUC$\uparrow$ & ACC$\uparrow$ \\
    \midrule
    & None 
    & 3.40 & 99.54 & 97.00 
    & 18.55$\pm$0.11 & 89.32$\pm$0.02 & 82.84$\pm$1.15 \\
    & 
    & & & 
    & (18.44, 18.66) & (89.34, 89.30) & (81.69, 83.98) \\
    & 1 
    & 2.85 & 99.58 & 97.11 
    & 19.79$\pm$2.55 & 88.81$\pm$2.89 & 82.93$\pm$0.11 \\
    & 
    & & & 
    & (17.24, 22.34) & (91.70, 85.92) & (82.82, 83.03) \\
    & 2 
    & 2.52 & 99.65 & 97.48 
    & 16.55$\pm$2.68 & 91.49$\pm$2.28 & 85.34$\pm$1.27 \\
    & 
    & & & 
    & (13.87, 19.23) & (93.77, 89.21) & (86.60, 84.07) \\
    \midrule
    \checkmark & None 
    & \underline{0.09} & \textbf{99.99} & \textbf{99.93} 
    & \underline{4.04$\pm$3.59} & 95.96$\pm$3.59 & 97.45$\pm$2.13 \\
    & 
    & & & 
    & (0.45, 7.63) & (99.55, 92.37) & (99.58, 95.32) \\
    \checkmark & 1 
    & 0.10 & \textbf{99.99} & 99.88 
    & 1.65$\pm$0.54 & \textbf{99.78$\pm$0.07} & \underline{98.49$\pm$0.44} \\
    & 
    & & & 
    & (1.11, 2.18) & (99.85, 99.71) & (98.93, 98.05) \\
    \checkmark & 2 
    & \textbf{0.07} & \textbf{99.99} & \underline{99.91} 
    & \textbf{0.63$\pm$0.18} & \underline{99.75$\pm$0.20} & \textbf{99.33$\pm$0.23} \\
    & 
    & & & 
    & (0.45, 0.81) & (99.55, 99.95) & (99.56, 99.10) \\
    \bottomrule
  \end{tabular}
\end{table*}
\begin{table*}[t]
  \centering
  \caption{Ablation study on the impact of the routing strategy. The best result is highlighted in \textbf{bold}. A checkmark (\checkmark) in ``Dynamic Routing'' indicates our PRN-based input-dependent prompt selection, while its absence denotes a static prompt configuration. The specific results for P2.1 and P2.2 are provided in parentheses below the average as (P2.1, P2.2). All metrics are reported as percentages [\%].}
  \label{tbl:Ablation_2}
  \setlength{\tabcolsep}{8pt}
  \begin{tabular}{@{}ccccccc@{}}
    \toprule
    \multirow{2}{*}{Dynamic Routing}
    & \multicolumn{3}{c}{P1} 
    & \multicolumn{3}{c}{P2} \\
    \cmidrule(lr){2-4} \cmidrule(lr){5-7}
    & ACER$\downarrow$ & AUC$\uparrow$ & ACC$\uparrow$ 
    & ACER$\downarrow$ & AUC$\uparrow$ & ACC$\uparrow$ \\
    \midrule
    & 0.08 & 99.99 & 99.87 
    & 5.23$\pm$4.84 & 98.25$\pm$1.74 & 94.85$\pm$4.77 \\
    & 
    & & & 
    (0.39, 10.07) & (99.99, 96.51) & (99.62, 90.08) \\
    \midrule
    \checkmark
    & \textbf{0.07} & \textbf{99.99} & \textbf{99.91} 
    & \textbf{0.63$\pm$0.18} & \textbf{99.75$\pm$0.20} & \textbf{99.33$\pm$0.23} \\
    & 
    & & & 
    (0.45, 0.81) & (99.55, 99.95) & (99.56, 99.10) \\
    \bottomrule
  \end{tabular}
\end{table*}

\paragraph{Ablation Study.}

To validate the core architectural contributions of our framework, we perform ablation studies on the hierarchical depth and routing strategy as shown in Tables \ref{tbl:Ablation} and \ref{tbl:Ablation_2}. 

First, we evaluate the impact of SSC and the VP-Tree depth.
As shown in Table \ref{tbl:Ablation}, the baseline without SSC fails under cross-modality conditions (ACER $>18\%$).
Applying SSC (Depth None) significantly improves performance but exhibits a significant asymmetry: it achieves 0.45\% ACER on P2.1 but degrades to 7.63\% on P2.2.
Introducing the hierarchical PRN (Depth 2) effectively resolves this imbalance, reducing the P2.2 ACER to 0.81\% and achieving the lowest average ACER of 0.63\%.
This result demonstrates that augmentation is insufficient to handle complex domain shifts; the hierarchical PRN is essential for bridging the gap across diverse attack modalities.

Table \ref{tbl:Ablation_2} confirms the necessity of input-dependent prompt selection.
We compare our dynamic routing with a static configuration where the same learnable prompts are fixed at each layer.
While the static approach performs well on seen attacks (P1), it overfits to specific modalities under cross-domain conditions, resulting in a catastrophic 10.07\% ACER on P2.2.
In contrast, our dynamic PRN activates the appropriate prompt subsets based on the input features, generalizing in both directions.
These results demonstrate that the dynamic routing mechanism of the VP-Tree is indispensable for robust unified attack detection.

\section{Conclusion}

In this paper, we proposed DINO-VPT, a lightweight, vision-only framework for unified physical-digital face anti-spoofing.
By integrating a hierarchical VP-Tree with dynamic routing, our method enables input-conditioned feature specialization while maintaining the computational efficiency of a single backbone.
Combined with SSC to mitigate modality imbalance, DINO-VPT achieves state-of-the-art cross-modality generalization on the challenging UniAttackData benchmark.
This approach successfully resolves the performance asymmetry between physical and digital domain shifts without relying on external vision-language models.
In future work, we will explore more expressive routing strategies, such as intermediate feature-based routing, to further enhance reliability in complex scenarios.
Additionally, we plan to extend this hierarchical prompt framework to datasets lacking explicit hierarchical annotations by uncovering latent label structures.

\section{Acknowledgment}

This work was supported in part by JSPS KAKENHI JP 23H00463 and 25K03131, and JST Moonshot R\&D Grant Number JPMJMS2215.

{\small
\bibliographystyle{ieee}
\bibliography{egbib}

\begin{thebibliography}{10}\itemsep=-1pt

\bibitem{Boulkenafet-FG-2017}
Z.~Boulkenafet, J.~Komulainen, L.~Li, X.~Feng, and A.~Hadid.
\newblock {OULU-NPU}: {A} mobile face presentation attack database with real-world variations.
\newblock {\em IEEE Int'l Conf. Automatic Face Gesture Recog.}, pages 612--618, June 2017.

\bibitem{Chen-IJCV-2026}
S.~Chen, A.~Liu, H.~Fang, H.~Yuan, J.~Zheng, D.~Zeng, Y.~Liu, J.~Deng, S.~Escalera, X.~Liu, J.~Wan, and Z.~Lei.
\newblock Uniattack: Unified physical-digital face attack detection.
\newblock {\em Int. J. Comput. Vis.}, 134:88, Jan. 2026.

\bibitem{Chen-AAAI-2025}
S.~Chen, A.~Liu, J.~Zheng, J.~Wan, K.~Peng, S.~Escalera, and Z.~Lei.
\newblock Mixture-of-attack-experts with class regularization for unified physical-digital face attack detection.
\newblock {\em AAAI}, 39:2195--2203, Apr. 2025.

\bibitem{IDIAP}
I.~Chingovska, A.~Anjos, and S.~Marcel.
\newblock On the effectiveness of local binary patterns in face anti-spoofing.
\newblock {\em Int. Conf. Biometrics Special Interest Group}, pages 1--7, Sept. 2012.

\bibitem{Dang-CVPR-2020}
H.~Dang, F.~Liu, J.~Stehouwer, X.~Liu, , and A.~K. Jain.
\newblock On the detection of digital face manipulation.
\newblock {\em IEEE/CVF Conf. Comput. Vis. Pattern Recog.}, pages 5781--5790, Oct. 2020.

\bibitem{Darcet-ICLR-2024}
T.~Darcet, M.~Oquab, J.~Mairal, and P.~Bojanowski.
\newblock Vision transformer needs register.
\newblock {\em Int. Conf. Learn. Represent.}, Apr. 2024.

\bibitem{Dosovitskiy-ICLR-2021}
A.~Dosovitskiy, L.~Beyer, A.~Kolesnikov, D.~Weissenborn, X.~Zhai, T.~Unterthiner, M.~Dehghani, M.~Minderer, G.~Heigold, S.~Gelly, J.~Uszkoreit, and N.~Houlsby.
\newblock An image is worth 16x16 words: {T}ransformers for image recognition at scale.
\newblock {\em Int. Conf. Learn. Represent.}, Jan. 2021.

\bibitem{Feng-ICCVW-2025}
M.~Feng, G.-M.~P. A., K.~Ito, and T.~Aoki.
\newblock Optimizing {DINOv2} with registers for face anti-spoofing.
\newblock {\em Int. Conf. Comput. Vis. Worksh.}, pages 3256--3262, Oct. 2025.

\bibitem{FENG-arXiv-2026}
M.~Feng, P.~Gallin-Martel, K.~Ito, and T.~Aoki.
\newblock Benchmarking vision foundation models for domain-generalizable face anti-spoofing.
\newblock {\em IEEE/CVF Conf. Comput. Vis. Pattern Recog. Worksh.}, Apr. 2026.

\bibitem{Feng-CVPRW-2025}
M.~Feng, K.~Ito, T.~Aoki, T.~Ohki, and M.~Nishigaki.
\newblock Leveraging intermediate features of vision transformer for face anti-spoofing.
\newblock {\em IEEE/CVF Conf. Comput. Vis. Pattern Recog. Worksh.}, pages 3464--3472, June 2025.

\bibitem{He-CVPR-2016}
K.~He, X.~Zhang, S.~Ren, and J.~Sun.
\newblock Deep residual learning for image recognition.
\newblock {\em IEEE/CVF Conf. Comput. Vis. Pattern Recog.}, pages 770--778, June 2016.

\bibitem{He-CVPRW-2024}
X.~He, D.~Liang, S.~Yang, Z.~Hao, H.~Ma, M.~Binjie, X.~Li, Y.~Wang, P.~Yan, and A.~Liu.
\newblock Joint physical-digital facial attack detection via simulating spoofing clues.
\newblock {\em IEEE/CVF Conf. Comput. Vis. Pattern Recog. Worksh.}, pages 995--1004, June 2024.

\bibitem{Jia-CVPR-2020}
Y.~Jia, J.~Zhang, S.~Shan, and X.~Chen.
\newblock Single-side domain generalization for face anti-spoofing.
\newblock {\em IEEE/CVF Conf. Comput. Vis. Pattern Recog.}, pages 8484--8493, June 2020.

\bibitem{Le-CVPR-2024}
B.~Le and S.~Woo.
\newblock Gradient alignment for cross-domain face anti-spoofing.
\newblock {\em IEEE/CVF Conf. Comput. Vis. Pattern Recog.}, pages 188--189, June 2024.

\bibitem{Li-NN-2024}
D.~Li, G.~Chen, X.~Wu, Z.~Yu, and M.~Tan.
\newblock Face anti-spoofing with cross-stage relation enhancement and spoof material perception.
\newblock {\em Neural Networks}, 175:106275, July 2024.

\bibitem{Handbook-Face-Recognition}
S.~Li and A.~Jain.
\newblock {\em Handbook of Face Recognition}.
\newblock Springer, 2011.

\bibitem{Liao-WACV-2023}
C.~Liao, W.~Chen, H.~Liu, Y.~Yeh, M.~Hu, and C.~Chen.
\newblock Domain invariant vision transformer learning for face anti-spoofing.
\newblock {\em IEEE/CVF Winter Conf. Applications of Comput. Vis.}, pages 6087--6096, Jan. 2023.

\bibitem{Liu-ICCV-2023}
A.~Liu, S.~Xue, J.~Gan, J.~Wan, L.~Y., J.~Deng, S.~Escalera, and Z.~Lei.
\newblock {CFPL-FAS}: {C}lass free prompt learning for generalizable face anti-spoofing.
\newblock {\em IEEE/CVF Conf. Comput. Vis. Pattern Recog.}, pages 222--232, 2024.

\bibitem{Liu-arXiv-2025}
A.~Liu, H.~Yuan, X.~Guo, H.~Ma, W.~Zhuang, Y.~Miao, C.~Hong, C.~Song, J.~Lan, Q.~Chu, T.~Gong, Y.~Liang, W.~Wang, J.~Wan, X.~Liu, and Z.~Lei.
\newblock Benchmarking unified face attack detection via hierarchical prompt tuning.
\newblock {\em arXiv preprint}, July 2025.

\bibitem{Liu-ACM-2021}
S.~Liu, K.~Zhang, T.~Yao, M.~Bi, S.~Ding, J.~Li, M.~Huang, and L.~Ma.
\newblock Adaptive normalized representation learning for generalizable face anti-spoofing.
\newblock {\em ACM Int. Conf. Multimedia}, pages 1469--1477, Oct. 2021.

\bibitem{Liu-IJCAI-2021}
S.~Liu, K.~Zhang, T.~Yao, K.~Sheng, S.~Ding, Y.~Tai, J.~Li, Y.~Xie, and L.~Ma.
\newblock Dual reweighting domain generalization for face presentation attack detection.
\newblock {\em IJCAI}, pages 867--873, 2021.

\bibitem{Liu-CVPR-2018}
Y.~Liu, A.~Jourabloo, and X.~Liu.
\newblock Learning deep models for face anti-spoofing: {B}inary or auxiliary supervision.
\newblock {\em IEEE/CVF Conf. Comput. Vis. Pattern Recog.}, pages 389--398, June 2018.

\bibitem{AdamW}
I.~Loshchilov and F.~Hutter.
\newblock Decoupled weight decay regularization.
\newblock Int. Conf. Learn. Represent., 2019.

\bibitem{Handbook-Anti-Spoofing}
S.~Marcel, J.~Fierrez, and N.~Evans.
\newblock {\em Handbook of Biometric Anti-Spoofing}.
\newblock Springer, 2023.

\bibitem{Radford-ICML-2021}
A.~Radford, J.~Kim, C.~Hallacy, A.~Ramesh, G.~Goh, S.~Agarwal, G.~Sastry, A.~Askell, P.~Mishkin, J.~Clark, G.~Krueger, and I.~Sutskever.
\newblock Learning transferable visual models from natural language supervision.
\newblock {\em Int. Conf. Learn. Represent.}, pages 8748--8763, July 2021.

\bibitem{Shao-CVPR-2019}
R.~Shao, X.~Lan, J.~Li, and P.~Yuen.
\newblock Multi-adversarial discriminative deep domain generalization for face presentation attack detection.
\newblock {\em IEEE/CVF Conf. Comput. Vis. Pattern Recog.}, pages 10023--10031, June 2019.

\bibitem{Srivatsan-ICCV-2023}
K.~Srivatsan, M.~Naseer, and K.~Nandakumar.
\newblock {FLIP}: {C}ross-domain face anti-spoofing with language guidance.
\newblock {\em Int. Conf. Comput. Vis.}, pages 19685--19696, Oct. 2023.

\bibitem{Wang-CVPR-2022}
C.~Wang, Y.~Lu, S.~Yang, and S.~Lai.
\newblock Patch{N}et: {A} simple face anti-spoofing framework via fine-grained patch recognition.
\newblock {\em IEEE/CVF Conf. Comput. Vis. Pattern Recog.}, pages 20281--20290, June 2022.

\bibitem{Wang-CVPR-2020}
G.~Wang, H.~Han, S.~Shan, and X.~Chen.
\newblock Cross-domain face presentation attack detection via multi-domain disentangled representation learning.
\newblock {\em IEEE/CVF Conf. Comput. Vis. Pattern Recog.}, pages 6677--6686, June 2020.

\bibitem{SSAN}
Z.~Wang, Z.~Wang, Z.~Yu, W.~Deng, J.~Li, T.~Gao, and Z.~Wang.
\newblock Domain generalization via shuffled style assembly for face anti-spoofing.
\newblock {\em IEEE/CVF Conf. Comput. Vis. Pattern Recog.}, pages 4123--4133, June 2022.

\bibitem{Wang-ECCV-2022}
Z.~Wang, Z.~Zhang, S.~Ebrahimi, R.~Sun, H.~Zhang, C.-Y. Lee, X.~Ren, G.~Su, V.~Perot, J.~Dy, and T.~Pfister.
\newblock {DualPrompt}: {C}omplementary prompting for rehearsal-free continual learning.
\newblock {\em Eur. Conf. Comput. Vis.}, page 631–648, Nov. 2022.

\bibitem{Wang-CVPR-2022-prompt}
Z.~Wang, Z.~Zhang, C.-Y. Lee, H.~Zhang, R.~Sun, and X.~Ren.
\newblock Learning to prompt for continual learning.
\newblock {\em IEEE/CVF Conf. Comput. Vis. Pattern Recog.}, pages 139--149, June 2022.

\bibitem{MSU-MFSD}
D.~Wen, A.~K. Jain, and H.~Han.
\newblock Face spoof detection with image distortion analysis.
\newblock {\em IEEE Trans. Inf. Forensics Secur.}, pages 746--761, Feb. 2015.

\bibitem{Xie-ICME-2025}
Z.~Xie, C.~Miao, A.~Liu, J.~Guo, F.~Li, D.~Guo, and Y.~Diao.
\newblock Suede:shared unified experts for physical-digital face attack detection enhancement.
\newblock {\em arXiv preprint}, Apr. 2025.

\bibitem{Yu-CVPR-2020}
Z.~Yu, C.~Zhao, Z.~Wang, Y.~Qin, Z.~Su, X.~Li, F.~Zhou, and G.~Zhao.
\newblock Searching central difference convolutional networks for face anti-spoofing.
\newblock {\em IEEE/CVF Conf. Comput. Vis. Pattern Recog.}, pages 5295--5305, June 2020.

\bibitem{Zhang-AAAI-2025}
G.~Zhang, K.~Wang, H.~Yue, A.~Liu, G.~Zhang, K.~Yao, E.~Ding, and J.~Wang.
\newblock Interpretable face anti-spoofing: {E}nhancing generalization with multimodal large language models.
\newblock {\em AAAI}, pages 9896--9904, Feb. 2025.

\bibitem{CASIA-FASD}
Z.~Zhang, J.~Yan, S.~Liu, Z.~Lei, D.~Yi, and S.~Z. Li.
\newblock A face antispoofing database with diverse attacks.
\newblock {\em Int. Conf. Biometrics}, pages 26--31, Mar. 2012.

\bibitem{Zhou-CVPR-2023}
Q.~Zhou, K.~Zhang, T.~Yao, X.~Lu, R.~Yi, S.~Ding, and L.~Ma.
\newblock Instance-aware domain generalization for face anti-spoofing.
\newblock {\em IEEE/CVF Conf. Comput. Vis. Pattern Recog.}, pages 20453--20463, June 2023.

\end{thebibliography}
}

\end{document}